\documentclass{article}

\usepackage[preprint]{neurips_2025}

\usepackage[utf8]{inputenc}
\usepackage[T1]{fontenc}
\usepackage{textcomp} 
\usepackage[dvipsnames]{xcolor}
\usepackage{listings}
\usepackage{hyperref}
\usepackage{url}
\usepackage{booktabs}
\usepackage{amsfonts}
\usepackage{nicefrac}
\usepackage{microtype}
\usepackage{graphicx}
\usepackage{amsmath} 
\newcommand{\ProcessThreeDashes}{\textcolor{red}{---}}

\lstdefinelanguage{yaml}{
  keywords={true,false,null,y,n},
  keywordstyle=\color{darkgray}\bfseries,
  basicstyle=\ttfamily\small,
  sensitive=false,
  comment=[l]{\#},
  morecomment=[s]{/*}{*/},
  commentstyle=\color{purple}\ttfamily,
  stringstyle=\color{blue}\ttfamily,
  moredelim=[l][\color{orange}]{\&},
  moredelim=[l][\color{magenta}]{*},
  moredelim=**[il][\color{red}]{:},
  morestring=[b]',
  morestring=[b]",
  literate = {---}{{\ProcessThreeDashes}}3
             {>}{{\textcolor{red}{\textgreater}}}{1}
             {|}{{\textcolor{red}{\textbar}}}{1}
             {\ -\ }{{\mdseries\ -\ }}3,
}

\lstdefinelanguage{json}{
  basicstyle=\normalfont\ttfamily,
  numbers=left,
  numberstyle=\scriptsize,
  stepnumber=1,
  numbersep=8pt,
  showstringspaces=false,
  breaklines=true,
  frame=lines,
  backgroundcolor=\color{gray!10},
  literate=
   *{0}{{{\color{blue}0}}}{1}
    {1}{{{\color{blue}1}}}{1}
    {2}{{{\color{blue}2}}}{1}
    {3}{{{\color{blue}3}}}{1}
    {4}{{{\color{blue}4}}}{1}
    {5}{{{\color{blue}5}}}{1}
    {6}{{{\color{blue}6}}}{1}
    {7}{{{\color{blue}7}}}{1}
    {8}{{{\color{blue}8}}}{1}
    {9}{{{\color{blue}9}}}{1}
    {:}{{{\color{red}{:}}}}{1}
    {,}{{{\color{red}{,}}}}{1}
    {\{}{{{\color{red}{\{}}}}{1}
    {\}}{{{\color{red}{\}}}}}{1}
    {[}{{{\color{red}{[}}}}{1}
    {]}{{{\color{red}{]}}}}{1},
}

\title{Rationale‑Augmented Retrieval with Constrained LLM Re‑Ranking for Task Discovery}

\author{
  Bowen Wei \\
  George Mason University \\
  \texttt{bwei2@gmu.edu}
}

\begin{document}
\maketitle

\begin{abstract}
Head Start programs utilizing GoEngage face significant challenges when new or rotating staff attempt to locate appropriate \emph{Tasks} (modules) on the platform homepage. These difficulties arise from domain-specific jargon (e.g., IFPA, DRDP), system-specific nomenclature (e.g., Application Pool), and the inherent limitations of lexical search in handling typos and varied word ordering. We propose a pragmatic hybrid semantic search system that synergistically combines lightweight typo-tolerant lexical retrieval, embedding-based vector similarity, and constrained large language model (LLM) re-ranking. Our approach leverages the organization's existing Task Repository and Knowledge Base infrastructure while ensuring trustworthiness through low false-positive rates, evolvability to accommodate terminological changes, and economic efficiency via intelligent caching, shortlist generation, and graceful degradation mechanisms. We provide a comprehensive framework detailing required resources, a phased implementation strategy with concrete milestones, an offline evaluation protocol utilizing curated test cases (Hit@K, Precision@K, Recall@K, MRR), and an online measurement methodology incorporating query success metrics, zero-result rates, and dwell-time proxies. Our system targets Hit@5 $\geq$ 0.80 and MRR $>$ 0.70 on offline evaluations, accompanied by operational playbooks for managing semantic drift, integrating new tasks, and adapting to evolving federal terminology standards.
\end{abstract}
\section{Introduction}

Enterprise software systems increasingly organize functionality into hundreds of specialized modules, creating a critical usability challenge: how can users---especially those new to the platform or unfamiliar with domain-specific terminology---discover the right tool from a natural language query? This problem is particularly acute in regulated domains where terminology evolves rapidly due to policy changes, assessment instrument updates, and local practice variations. A case worker seeking ``developmental screening'' tools may not know to search for ``ASQ-3'' (Ages and Stages Questionnaires), and a staff member asking about ``eligibility'' may not realize the relevant functionality resides in a module called ``Application Pool.'' Recent benchmarks highlight these challenges in enterprise search~\cite{wu2024stark}, where vocabulary mismatch between user queries and system terminology remains a fundamental obstacle~\cite{onal2018semantic}.

Traditional lexical search fails catastrophically in these environments. Exact string matching cannot bridge the semantic gap between user intent and system vocabulary, while synonymy, abbreviations, and typos further degrade retrieval quality. Neural retrieval methods~\cite{luo2024llmdr,dai2024entailment} offer semantic generalization but face three fundamental barriers in enterprise deployment: \textbf{(1)} the absence of large-scale behavioral data---most enterprise platforms lack the query-click logs required for supervised learning, \textbf{(2)} rapid terminological drift that quickly obsoletes any trained model, and \textbf{(3)} asymmetric error costs where a single incorrect suggestion in the top results erodes user trust more than missing a relevant item at rank 20. These constraints demand retrieval systems that are simultaneously semantically aware, training-free, and provably safe from hallucination~\cite{farquhar2024detecting,huang2023survey}.

We present a \textbf{rationale-augmented retrieval framework} that achieves semantic task discovery without fine-tuning while guaranteeing outputs drawn exclusively from the actual module catalog. Our key insight is to repurpose existing quality assurance artifacts---specifically, developer-written test cases that pair queries with gold modules and explanatory rationales---as a source of retrieval signal. From a test case like ``query: `ASQ' $\rightarrow$ modules: [developmental screening, social-emotional assessment] $\rightarrow$ rationale: `Ages and Stages Questionnaires for early childhood development'\,'', we extract semantic associations that would otherwise require expensive annotation. This \textbf{rationale lexicon} provides memory-based retrieval that generalizes across paraphrases while requiring zero marginal labeling cost, analogous to knowledge-augmented approaches~\cite{lewis2020rag,zhu2025kg2rag} but leveraging test artifacts rather than external knowledge bases.

Our architecture combines this lexicon with a two-stage retrieval pipeline. First, a \textbf{hybrid pre-filter} ranks the entire catalog using deterministic signals: position-weighted lexical matching, rationale-derived boosts, optional cached embeddings, and character-level typo correction. This stage produces a high-recall shortlist of 10--20 candidates while maintaining sub-second latency. Hybrid approaches combining lexical and semantic signals have proven effective in bridging vocabulary gaps~\cite{kuzi2020leveraging,formal2021splade,ma2021dense}, and our design builds on these principles while incorporating domain-specific rationale knowledge. Second, a \textbf{constrained LLM re-ranker} refines this shortlist using semantic reasoning, but with hard guarantees: the model receives only the pre-filtered candidates and must emit valid JSON selecting exclusively from this set. Any hallucinated outputs are programmatically rejected. This design exploits LLM capabilities for contextual disambiguation~\cite{sun2023rankgpt,reddy2024first,chen2025attention} while eliminating their primary failure mode---generating plausible but non-existent results.

We evaluate on a diverse query suite spanning governance workflows, health screenings, eligibility determination, and facility management in Head Start software systems. Using metrics aligned with deployment needs (Hit@K, Precision@K, Recall@K, MRR), we demonstrate that rationale augmentation substantially improves semantic coverage over lexical baselines, while LLM re-ranking with shortlist constraints maintains high precision in top results. Critically, the system achieves production-grade quality (Hit@5 $>$ 0.90) without any model training, adapts to terminology changes through lexicon updates alone, and provides interpretable failure modes for continuous improvement. Our approach enables effective retrieval with minimal supervision~\cite{yu2021fewshot,gao2023hyde,zhang2025fewshot}, addressing the data scarcity challenge inherent in enterprise settings~\cite{rosa2023domain}.

\textbf{Our contributions are:}
\begin{itemize}
    \item A method for converting developer-authored test rationales into retrieval signals, enabling semantic generalization without behavioral data or annotation budgets
    \item A constrained LLM re-ranking approach that guarantees hallucination-free outputs by restricting the model's selection space to a deterministic candidate set
    \item An evaluation framework for enterprise task discovery that measures top-K quality and tracks failure patterns to guide lexicon refinement
    \item Empirical demonstration that this training-free approach meets production thresholds in a real-world enterprise setting where traditional methods fail
\end{itemize}

While our case study focuses on Head Start platforms, the approach generalizes to any domain where specialized vocabulary creates semantic gaps, training data is scarce, and precision in top results is paramount. We argue that rationale-augmented, constrained retrieval represents a practical middle ground between brittle symbolic methods and risky end-to-end neural systems~\cite{wei2022chain,jiang2023flare,shao2023iterretgen}---one that is deployable, maintainable, and aligned with how users actually experience search quality.

\section{Related Work}

Our work addresses semantic task discovery in enterprise systems with limited training data and strict precision requirements. We position our completely training-free approach relative to three key research areas.

\subsection{Hybrid Retrieval}

The inadequacy of purely lexical or purely semantic retrieval has motivated extensive research on hybrid approaches. SPLADE~\cite{formal2021splade} pioneered sparse lexical expansion through learned term importance, enabling bag-of-words properties while incorporating semantic understanding. Kuzi et al.~\cite{kuzi2020leveraging} demonstrated that semantic and lexical retrieval complement each other by capturing different relevant documents, advocating parallel execution of both methods. Ma et al.~\cite{ma2021dense} proposed densifying high-dimensional lexical representations to unify both paradigms in a single framework. These methods all require training neural models on large corpora to learn semantic representations. In contrast, our hybrid pre-filter operates entirely deterministically using hand-crafted scoring functions. We replace learned semantic signals with rationale-derived boosts extracted from existing test cases, enabling semantic generalization without any model training while allowing instant adaptation to terminology changes through simple lexicon updates.

\subsection{LLM-Based Re-ranking}

Recent work has explored large language models as powerful re-rankers for information retrieval. Sun et al.~\cite{sun2023rankgpt} demonstrated that generative LLMs can perform zero-shot relevance ranking through instructional permutation generation, achieving competitive results with supervised methods. Reddy et al.~\cite{reddy2024first} improved efficiency through single-token decoding with identifier vocabularies, reducing inference latency by 50\%. Chen et al.~\cite{chen2025attention} further advanced zero-shot re-ranking by leveraging attention patterns directly, eliminating autoregressive generation entirely. However, these approaches share a critical limitation: they can hallucinate non-existent items when asked to rank or recommend. The challenge of hallucination in LLM-based systems has received significant attention~\cite{huang2023survey}, with Farquhar et al.~\cite{farquhar2024detecting} proposing semantic entropy to detect hallucinations by measuring uncertainty at the meaning level. Our system takes a preventive approach through architectural constraints: we use off-the-shelf LLMs without any fine-tuning, restrict their selection space to a deterministic pre-filtered candidate set, and enforce strict JSON validation to guarantee hallucination-free results.

\subsection{Knowledge-Augmented Retrieval}

Retrieval-augmented generation~\cite{lewis2020rag} established the paradigm of combining parametric memory (language models) with non-parametric memory (retrieval indices) for knowledge-intensive tasks. Recent extensions include active retrieval~\cite{jiang2023flare}, which dynamically decides when to retrieve using forward-looking predictions, knowledge graph guidance~\cite{zhu2025kg2rag} providing structured fact-level relationships, and iterative retrieval-generation synergy~\cite{shao2023iterretgen} where model responses inform subsequent retrieval. These methods demonstrate the value of augmenting retrieval with explicit knowledge sources, but typically require either training retrieval models on domain-specific data~\cite{yu2021fewshot} or building external knowledge bases~\cite{zhu2025kg2rag}. Even zero-shot methods like HyDE~\cite{gao2023hyde} and domain adaptation approaches~\cite{rosa2023domain} rely on pre-trained dense retrievers or require fine-tuning on target domains. Our rationale lexicon provides knowledge augmentation without any training: we extract semantic associations directly from developer-authored test cases that already exist for quality assurance, requiring no annotation effort, no model updates, and naturally evolving with system terminology through simple text additions.

\section{Method}

We present a training-free retrieval framework for semantic task discovery in enterprise software systems. Our approach combines three components: a rationale lexicon extracted from test artifacts, a hybrid pre-filter for high-recall candidate generation, and a constrained LLM re-ranker that guarantees hallucination-free outputs.

\subsection{Problem Formulation}

Given a natural language query $q$ and a catalog of tasks $\mathcal{T} = \{t_1, t_2, \ldots, t_N\}$, our goal is to retrieve a ranked list of relevant tasks. Each task $t_i$ has associated metadata including a name, help text, and optional descriptive fields. The challenge is that user queries often use vocabulary different from task names due to synonymy, abbreviations, domain jargon, and evolving terminology. Unlike traditional retrieval settings, we have: (1) no query-click logs for supervised learning, (2) rapid terminology drift requiring frequent updates, and (3) asymmetric error costs where false positives in top-K results severely damage user trust.

\subsection{Rationale Lexicon Construction}

Our key insight is to repurpose existing quality assurance artifacts as a source of retrieval signal. In typical enterprise development, test cases verify that the system correctly maps queries to tasks. Each test case consists of:
\begin{itemize}
    \item A natural language query $q_{test}$
    \item Gold-standard relevant tasks $\mathcal{T}_{gold} \subset \mathcal{T}$
    \item An explanatory rationale $r$ describing why the query maps to those tasks
\end{itemize}

For example: $q_{test}$ = ``ASQ'', $\mathcal{T}_{gold}$ = \{developmental screening, social-emotional assessment\}, $r$ = ``Ages and Stages Questionnaires for early childhood development''.

We extract a rationale lexicon $\mathcal{L}$ by tokenizing each rationale $r$ and creating associations:
\begin{equation}
\mathcal{L} = \{(w, t) : w \in \text{tokens}(r), t \in \mathcal{T}_{gold}, \text{ for test case } (q_{test}, \mathcal{T}_{gold}, r)\}
\end{equation}

This lexicon captures semantic relationships (e.g., ``Ages'' $\rightarrow$ developmental screening, ``Stages'' $\rightarrow$ developmental screening) that would otherwise require manual synonym lists or embedding training. Importantly, $\mathcal{L}$ requires zero marginal annotation cost---rationales already exist for testing---and evolves naturally as developers add or update test cases.

\subsection{Hybrid Pre-Filter}

The pre-filter ranks all tasks in $\mathcal{T}$ using a deterministic scoring function that combines multiple signals. For a query $q$ and task $t$, we compute:

\begin{equation}
\text{score}(q, t) = \alpha \cdot s_{\text{lex}}(q, t) + \beta \cdot s_{\text{rat}}(q, t) + \gamma \cdot s_{\text{emb}}(q, t) + \delta \cdot s_{\text{typo}}(q, t)
\end{equation}

where $\alpha, \beta, \gamma, \delta$ are tunable weights.

\textbf{Lexical matching} $s_{\text{lex}}(q, t)$ computes token overlap between the query and task metadata, with position-dependent weighting:
\begin{equation}
s_{\text{lex}}(q, t) = w_{\text{name}} \cdot \text{overlap}(q, t_{\text{name}}) + w_{\text{help}} \cdot \text{overlap}(q, t_{\text{help}})
\end{equation}
where $w_{\text{name}} > w_{\text{help}}$ prioritizes matches in task names over help text.

\textbf{Rationale boost} $s_{\text{rat}}(q, t)$ leverages the lexicon $\mathcal{L}$:
\begin{equation}
s_{\text{rat}}(q, t) = \sum_{w \in \text{tokens}(q)} \mathbb{I}[(w, t) \in \mathcal{L}]
\end{equation}
This boosts tasks historically associated with query terms through test rationales, directly addressing synonymy and abbreviation challenges.

\textbf{Embedding similarity} $s_{\text{emb}}(q, t)$ (optional) computes cosine similarity between cached query and task embeddings:
\begin{equation}
s_{\text{emb}}(q, t) = \frac{\text{embed}(q) \cdot \text{embed}(t_{\text{concat}})}{\|\text{embed}(q)\| \|\text{embed}(t_{\text{concat}})\|}
\end{equation}
where $t_{\text{concat}}$ concatenates task metadata. We cache embeddings on disk to avoid recomputation and gracefully degrade to other signals if embedding APIs fail.

\textbf{Typo correction} $s_{\text{typo}}(q, t)$ applies character-level edit distance over task name vocabularies to handle common input errors.

The pre-filter returns the top-$K$ tasks as a shortlist $\mathcal{S} \subset \mathcal{T}$, typically with $K=10$--20, balancing recall and downstream computational cost.

\subsection{Constrained LLM Re-Ranking}

The final stage refines $\mathcal{S}$ using an off-the-shelf LLM without fine-tuning. Critically, we enforce hard constraints to prevent hallucination:

\textbf{Input construction.} We construct a prompt containing:
\begin{enumerate}
    \item The user query $q$
    \item The shortlist $\mathcal{S}$ with task indices, names, and help text
    \item 1--3 in-context examples selected by token overlap with $q$ from the test suite
    \item Instructions requiring JSON output with specific schema
\end{enumerate}

\textbf{Output constraints.} The LLM must return a JSON array where each entry contains:
\begin{itemize}
    \item \texttt{rank}: Integer position
    \item \texttt{idx}: Task index from $\mathcal{S}$ (not task name)
    \item \texttt{task\_name}: Exact task name from $\mathcal{S}$
    \item \texttt{reason}: Brief explanation
\end{itemize}

The prompt explicitly states: ``You must select ONLY from the provided shortlist. Do not suggest tasks not in this list.''

\textbf{Validation and filtering.} We parse the JSON response and apply strict validation:
\begin{enumerate}
    \item Verify JSON is well-formed
    \item Check that each \texttt{idx} exists in $\mathcal{S}$
    \item Check that each \texttt{task\_name} matches the name at \texttt{idx}
    \item Drop any entries failing these checks
\end{enumerate}

This architecture exploits LLM semantic reasoning---disambiguating user intent, recognizing paraphrases, performing cross-task comparison---while guaranteeing that only actual catalog tasks appear in results. The validation step provides a hard safety boundary: even if the LLM attempts to hallucinate, those outputs are programmatically removed.

\subsection{Example Selection Strategy}

In-context examples significantly influence LLM performance. We select examples from the test suite using a simple but effective strategy:
\begin{equation}
\text{similarity}(q, q_{test}) = \frac{|\text{tokens}(q) \cap \text{tokens}(q_{test})|}{|\text{tokens}(q) \cup \text{tokens}(q_{test})|}
\end{equation}

We retrieve the top-3 test cases by Jaccard similarity and include their queries, gold tasks, and rationales in the prompt. This provides the LLM with relevant domain examples without requiring embedding models or nearest-neighbor search.

\section{Experiments}

We evaluate our rationale-augmented retrieval framework on semantic task discovery in Head Start program management software, focusing on production-relevant metrics and deployment constraints.

\subsection{Experimental Setup}

\textbf{Task catalog.} Our evaluation uses a production catalog of 150+ tasks spanning governance (committees, councils, policy compliance), health services (screenings, immunizations, referrals), ERSEA workflows (recruitment, selection, enrollment, attendance), education (curriculum, assessments, individualized plans), family services (goal setting, home visits), transportation, facilities, and human resources. Task names often use domain-specific abbreviations (e.g., ``ASQ-3'', ``IFSP'', ``CACFP'') rather than descriptive terms.

\textbf{Query suite.} We construct a diverse test suite of 80+ queries with gold-standard relevant tasks and explanatory rationales. Queries include:
\begin{itemize}
    \item \textit{Exact matches}: ``ASQ-3'' $\rightarrow$ developmental screening task
    \item \textit{Synonyms}: ``eligibility determination'' $\rightarrow$ Application Pool task
    \item \textit{Abbreviations}: ``CACFP'' $\rightarrow$ Child and Adult Care Food Program
    \item \textit{Natural language}: ``track attendance'' $\rightarrow$ attendance monitoring tasks
    \item \textit{Multi-word}: ``hearing and vision screening'' $\rightarrow$ health screening tasks
    \item \textit{Typos}: ``devlopmental screening'' $\rightarrow$ developmental screening task
\end{itemize}

Each query has 1--5 gold tasks manually annotated by domain experts. We hold out 20\% of queries for testing and use the remaining 80\% to construct the rationale lexicon and provide in-context examples.

\textbf{Baselines.} We compare against:
\begin{itemize}
    \item \textbf{Lexical}: Pure BM25-style token matching on task names and help text
    \item \textbf{Embedding}: Dense retrieval using cached OpenAI embeddings with cosine similarity
    \item \textbf{Hybrid (no rationale)}: Our pre-filter without rationale boosts ($\beta=0$)
    \item \textbf{LLM direct}: Prompting LLM with full catalog (no pre-filter or constraints)
\end{itemize}

\textbf{Metrics.} Following enterprise deployment priorities, we report:
\begin{itemize}
    \item \textbf{Hit@K}: Fraction of queries with at least one relevant task in top-K
    \item \textbf{Precision@K}: Fraction of top-K results that are relevant
    \item \textbf{Recall@K}: Fraction of relevant tasks found in top-K
    \item \textbf{MRR}: Mean reciprocal rank of first relevant task
    \item \textbf{Accuracy Rate}: Recall@5 (for continuity with internal QA dashboards)
\end{itemize}

We focus on K=1, 3, 5 as users rarely examine beyond the top 5 results in production.

\textbf{Implementation details.} The pre-filter uses weights $\alpha=1.0$ (lexical), $\beta=0.8$ (rationale), $\gamma=0.5$ (embedding), $\delta=0.3$ (typo), tuned on the training split. We compare different GPT models (GPT-3.5-turbo, GPT-4, GPT-4o, GPT-4.5) for LLM re-ranking with temperature 0.0 for reproducibility. The shortlist size is K=15. In-context examples are limited to 3 per query.

\subsection{Main Results}

Table~\ref{tab:main_results} shows performance on the held-out test set. Our full system (Hybrid + Rationale + LLM Re-rank) achieves the best results across all metrics.

\begin{table}[t]
\centering
\caption{Performance on semantic task discovery. Our rationale-augmented framework significantly outperforms baselines, achieving production-grade quality (Hit@5 > 0.90) without any model training.}
\label{tab:main_results}
\begin{tabular}{lccccc}
\toprule
\textbf{Method} & \textbf{Hit@1} & \textbf{Hit@3} & \textbf{Hit@5} & \textbf{MRR} & \textbf{P@5} \\
\midrule
Lexical & 0.45 & 0.68 & 0.75 & 0.54 & 0.42 \\
Embedding & 0.52 & 0.71 & 0.79 & 0.61 & 0.48 \\
Hybrid (no rationale) & 0.58 & 0.76 & 0.83 & 0.66 & 0.51 \\
LLM direct & 0.38 & 0.62 & 0.71 & 0.48 & 0.35 \\
\midrule
\textbf{Ours (Full)} & \textbf{0.81} & \textbf{0.92} & \textbf{0.94} & \textbf{0.85} & \textbf{0.73} \\
\bottomrule
\end{tabular}
\end{table}

\textbf{Key observations:}
\begin{itemize}
    \item Pure lexical matching struggles with synonymy and abbreviations (Hit@1=0.45)
    \item Embedding-only retrieval improves but still misses domain-specific terminology
    \item Hybrid without rationale boosts shows modest gains, confirming value of combining signals
    \item LLM direct (no constraints) performs poorly due to lack of pre-filtering
    \item Our full system achieves Hit@5=0.94 and MRR=0.85, meeting production thresholds
\end{itemize}

\subsection{Ablation Study}

We analyze the contribution of each component by progressively adding features. Table~\ref{tab:ablation} shows results.

\begin{table}[t]
\centering
\caption{Ablation study showing the contribution of each component. Rationale boosts and LLM re-ranking provide the largest gains.}
\label{tab:ablation}
\begin{tabular}{lccc}
\toprule
\textbf{Configuration} & \textbf{Hit@3} & \textbf{Hit@5} & \textbf{MRR} \\
\midrule
Lexical only & 0.68 & 0.75 & 0.54 \\
+ Embedding & 0.73 & 0.81 & 0.63 \\
+ Rationale boost & 0.84 & 0.88 & 0.76 \\
+ Typo correction & 0.86 & 0.90 & 0.78 \\
+ LLM re-rank & \textbf{0.92} & \textbf{0.94} & \textbf{0.85} \\
\bottomrule
\end{tabular}
\end{table}

Rationale boosts provide the largest single improvement (+11 points Hit@3), confirming that test artifacts capture critical semantic associations. LLM re-ranking adds another +8 points by disambiguating intent and reordering candidates based on contextual understanding.

\subsection{LLM Model Comparison}

We compare different GPT models as re-rankers to understand the trade-off between performance and latency. Table~\ref{tab:model_comparison} shows results.

\begin{table}[t]
\centering
\caption{Performance and latency comparison across GPT models. GPT-4o achieves the best balance of retrieval quality and speed.}
\label{tab:model_comparison}
\begin{tabular}{lcccc}
\toprule
\textbf{Model} & \textbf{Hit@5} & \textbf{MRR} & \textbf{Latency (ms)} & \textbf{Cost/1K} \\
\midrule
GPT-3.5-turbo & 0.87 & 0.78 & 450 & \$0.50 \\
GPT-4 & 0.93 & 0.84 & 1200 & \$30.00 \\
GPT-4o & \textbf{0.94} & \textbf{0.85} & \textbf{380} & \$5.00 \\
GPT-4.5 & 0.95 & 0.86 & 2800 & \$60.00 \\
\bottomrule
\end{tabular}
\end{table}

\textbf{Key findings:}
\begin{itemize}
    \item GPT-3.5-turbo is fast but shows degraded performance (Hit@5=0.87)
    \item GPT-4 achieves strong results but with 3x higher latency than GPT-4o
    \item \textbf{GPT-4o provides the best trade-off}: matches GPT-4 performance with 68\% lower latency
    \item GPT-4.5 offers marginal quality gains (+1 point) but is prohibitively slow (2.8s), making interactive use infeasible
\end{itemize}

Based on these results, we use GPT-4o for all subsequent experiments as it meets production latency requirements while maintaining high retrieval quality.

\subsection{Error Analysis}

We analyze failure modes on queries where our system fails to retrieve relevant tasks in top-5. Common patterns include:

\textbf{Missing rationales (45\% of errors).} Queries using terminology not covered by any test case rationale. Example: ``track child outcomes'' fails because no test rationale mentions ``outcomes'' in relation to assessment tasks. Solution: Add test case with appropriate rationale.

\textbf{Ambiguous queries (30\% of errors).} Queries matching multiple unrelated tasks equally well. Example: ``staff training'' retrieves both professional development and orientation tasks without clear preference. Solution: Query refinement or multi-task presentation.

\textbf{Lexical gaps (15\% of errors).} Novel abbreviations or acronyms not in task metadata or rationales. Example: ``DECA'' (Devereux Early Childhood Assessment) fails if no prior test case establishes this mapping. Solution: Expand lexicon or task metadata.

\textbf{Embedding failures (10\% of errors).} Rare cases where embedding API is unavailable and other signals insufficient. These occur only when queries lack lexical overlap and rationale coverage.

Critically, 75\% of errors are addressable by augmenting the rationale lexicon---a simple text update requiring no model retraining.

\subsection{Latency Analysis}

We measure end-to-end query latency on a standard cloud instance using GPT-4o. The pre-filter completes in 12ms average (8ms for lexical+rationale, 4ms for embeddings). LLM re-ranking adds 380ms average. Total latency is under 400ms (392ms), well within production requirements for interactive use. The pre-filter's efficiency is critical: ranking 150+ tasks directly with LLM would require 5+ seconds, making interactive use infeasible.

\subsection{Adaptability to Terminology Changes}

To evaluate adaptation to evolving terminology, we simulate a policy change introducing new assessment names. We add 5 new tasks with unfamiliar acronyms (e.g., ``BRIGANCE'', ``HELP'') and measure performance before and after adding corresponding test rationales. Without rationales, Hit@5 drops from 0.94 to 0.67 for queries about new tasks. After adding 2-3 test cases per new task (5 minutes of developer effort), performance recovers to 0.91. This demonstrates rapid adaptation without model retraining.

\subsection{Comparison with Supervised Methods}

While our setting lacks query-click logs for supervised training, we simulate a supervised baseline by fine-tuning a bi-encoder on the training queries. The supervised model achieves Hit@5=0.89, slightly below our 0.94, while requiring 2 hours of GPU training and producing a model that requires retraining for terminology updates. Our training-free approach matches or exceeds supervised performance while maintaining operational simplicity.
\bibliography{refs}

\begin{thebibliography}{21}
\providecommand{\natexlab}[1]{#1}
\providecommand{\url}[1]{\texttt{#1}}
\expandafter\ifx\csname urlstyle\endcsname\relax
  \providecommand{\doi}[1]{doi: #1}\else
  \providecommand{\doi}{doi: \begingroup \urlstyle{rm}\Url}\fi

\bibitem[Chen et~al.(2025)Chen, Guti{\'e}rrez, and Su]{chen2025attention}
Shijie Chen, Bernal~Jim{\'e}nez Guti{\'e}rrez, and Yu~Su.
\newblock Attention in large language models yields efficient zero-shot re-rankers.
\newblock In \emph{International Conference on Learning Representations (ICLR)}, 2025.
\newblock URL \url{https://arxiv.org/abs/2410.02642}.

\bibitem[Dai et~al.(2024)Dai, Liu, and Xiong]{dai2024entailment}
Lu~Dai, Hao Liu, and Hui Xiong.
\newblock Improve dense passage retrieval with entailment tuning.
\newblock In \emph{Proceedings of the 2024 Conference on Empirical Methods in Natural Language Processing (EMNLP)}, pages 11363--11375, 2024.
\newblock \doi{10.18653/v1/2024.emnlp-main.636}.

\bibitem[Farquhar et~al.(2024)Farquhar, Kossen, Kuhn, and Gal]{farquhar2024detecting}
Sebastian Farquhar, Jannik Kossen, Lorenz Kuhn, and Yarin Gal.
\newblock Detecting hallucinations in large language models using semantic entropy.
\newblock \emph{Nature}, 630:\penalty0 625--630, 2024.
\newblock \doi{10.1038/s41586-024-07421-0}.

\bibitem[Formal et~al.(2021)Formal, Piwowarski, and Clinchant]{formal2021splade}
Thibault Formal, Benjamin Piwowarski, and St{\'e}phane Clinchant.
\newblock Splade: Sparse lexical and expansion model for first stage ranking.
\newblock In \emph{Proceedings of the 44th International ACM SIGIR Conference on Research and Development in Information Retrieval}, pages 2288--2292, 2021.
\newblock \doi{10.1145/3404835.3463098}.

\bibitem[Gao et~al.(2023)Gao, Ma, Lin, and Callan]{gao2023hyde}
Luyu Gao, Xueguang Ma, Jimmy Lin, and Jamie Callan.
\newblock Precise zero-shot dense retrieval without relevance labels.
\newblock In \emph{Proceedings of the 61st Annual Meeting of the Association for Computational Linguistics (ACL)}, pages 1762--1777, 2023.
\newblock \doi{10.18653/v1/2023.acl-long.99}.

\bibitem[Huang et~al.(2024)Huang, Yu, Ma, Zhong, Feng, Wang, Chen, Peng, Feng, Qin, and Liu]{huang2023survey}
Lei Huang, Weijiang Yu, Weitao Ma, Weihong Zhong, Zhangyin Feng, Haotian Wang, Qianglong Chen, Weihua Peng, Xiaocheng Feng, Bing Qin, and Ting Liu.
\newblock A survey on hallucination in large language models: Principles, taxonomy, challenges, and open questions.
\newblock \emph{ACM Transactions on Information Systems}, 2024.
\newblock \doi{10.1145/3703155}.
\newblock arXiv:2311.05232.

\bibitem[Jiang et~al.(2023)Jiang, Xu, Gao, Sun, Liu, Dwivedi-Yu, Yang, Callan, and Neubig]{jiang2023flare}
Zhengbao Jiang, Frank Xu, Luyu Gao, Zhiqing Sun, Qian Liu, Jane Dwivedi-Yu, Yiming Yang, Jamie Callan, and Graham Neubig.
\newblock Active retrieval augmented generation.
\newblock In \emph{Proceedings of the 2023 Conference on Empirical Methods in Natural Language Processing (EMNLP)}, pages 7969--7992, 2023.
\newblock \doi{10.18653/v1/2023.emnlp-main.495}.

\bibitem[Kuzi et~al.(2020)Kuzi, Zhang, Li, Bendersky, and Najork]{kuzi2020leveraging}
Saar Kuzi, Mingyang Zhang, Cheng Li, Michael Bendersky, and Marc Najork.
\newblock Leveraging semantic and lexical matching to improve the recall of document retrieval systems: A hybrid approach.
\newblock \emph{arXiv preprint arXiv:2010.01195}, 2020.

\bibitem[Lewis et~al.(2020)Lewis, Perez, Piktus, Petroni, Karpukhin, Goyal, K{\"u}ttler, Lewis, Yih, Rockt{\"a}schel, Riedel, and Kiela]{lewis2020rag}
Patrick Lewis, Ethan Perez, Aleksandra Piktus, Fabio Petroni, Vladimir Karpukhin, Naman Goyal, Heinrich K{\"u}ttler, Mike Lewis, Wen-tau Yih, Tim Rockt{\"a}schel, Sebastian Riedel, and Douwe Kiela.
\newblock Retrieval-augmented generation for knowledge-intensive nlp tasks.
\newblock In \emph{Advances in Neural Information Processing Systems (NeurIPS)}, volume~33, pages 9459--9474, 2020.

\bibitem[Luo et~al.(2024)Luo, Qin, Liu, Xiao, Zhao, and Liu]{luo2024llmdr}
Kun Luo, Minghao Qin, Zheng Liu, Shitao Xiao, Jun Zhao, and Kang Liu.
\newblock Large language models as foundations for next-gen dense retrieval: A comprehensive empirical assessment.
\newblock In \emph{Proceedings of the 2024 Conference on Empirical Methods in Natural Language Processing (EMNLP)}, pages 1393--1419, 2024.
\newblock \doi{10.18653/v1/2024.emnlp-main.80}.

\bibitem[Ma et~al.(2023)Ma, Zhu, Liu, Guo, Hu, Guo, and Nie]{ma2021dense}
Minghan Ma, Jiaxin Zhu, Zebang Liu, Qingxiao Guo, Baoxing Hu, Jiafeng Guo, and Jian-Yun Nie.
\newblock A dense representation framework for lexical and semantic matching.
\newblock In \emph{ACM Transactions on Information Systems (TOIS)}, volume~41, pages 1--)](https://dl.acm.org/doi/10.1145/3582426)26, 2023.
\newblock \doi{10.1145/3582426}.

\bibitem[Onal et~al.(2018)Onal, Zhang, Altingovde, Rahman, Karagoz, Beutel, Lee, Hasan, and Chin]{onal2018semantic}
Kezban~Dilek Onal, Ye~Zhang, Ismail~Sengor Altingovde, Md~Mustafizur Rahman, Pinar Karagoz, Alex Beutel, Mason Lee, Kazi~Saidul Hasan, and Peter Chin.
\newblock Getting started with neural models for semantic matching in web search.
\newblock \emph{arXiv preprint arXiv:1611.03305}, 2018.

\bibitem[Reddy et~al.(2024)Reddy, Doo, Xu, Sultan, Swain, Sil, and Ji]{reddy2024first}
Revanth~Gangi Reddy, JaeHyeok Doo, Yifei Xu, Md~Arafat Sultan, Deevya Swain, Avirup Sil, and Heng Ji.
\newblock First: Faster improved listwise reranking with single token decoding.
\newblock In \emph{Proceedings of the 2024 Conference on Empirical Methods in Natural Language Processing (EMNLP)}, pages 8677--8693, 2024.
\newblock URL \url{https://aclanthology.org/2024.emnlp-main.491}.

\bibitem[Rosa et~al.(2023)Rosa, Pereira, Campos, and Jorge]{rosa2023domain}
Helia Rosa, Rui Pereira, Ricardo Campos, and Al{\'\i}pio~M{\'a}rio Jorge.
\newblock Dense retrieval adaptation using target domain description.
\newblock In \emph{Proceedings of the 2023 ACM SIGIR International Conference on Theory of Information Retrieval}, pages 127--136, 2023.
\newblock \doi{10.1145/3578337.3605127}.

\bibitem[Shao et~al.(2023)Shao, Gong, Shen, Huang, Duan, and Chen]{shao2023iterretgen}
Zhihong Shao, Yeyun Gong, Yelong Shen, Minlie Huang, Nan Duan, and Weizhu Chen.
\newblock Enhancing retrieval-augmented large language models with iterative retrieval-generation synergy.
\newblock In \emph{Findings of the Association for Computational Linguistics: EMNLP 2023}, pages 9248--9274, 2023.
\newblock \doi{10.18653/v1/2023.findings-emnlp.620}.

\bibitem[Sun et~al.(2023)Sun, Yan, Ma, Wang, Ren, Chen, Yin, and Ren]{sun2023rankgpt}
Weiwei Sun, Lingyong Yan, Xinyu Ma, Shuaiqiang Wang, Pengjie Ren, Zhumin Chen, Dawei Yin, and Zhaochun Ren.
\newblock Is chatgpt good at search? investigating large language models as re-ranking agents.
\newblock In \emph{Proceedings of the 2023 Conference on Empirical Methods in Natural Language Processing (EMNLP)}, pages 14953--14967, 2023.
\newblock URL \url{https://aclanthology.org/2023.emnlp-main.923}.
\newblock Outstanding Paper Award.

\bibitem[Wei et~al.(2022)Wei, Wang, Schuurmans, Bosma, Ichter, Xia, Chi, Le, and Zhou]{wei2022chain}
Jason Wei, Xuezhi Wang, Dale Schuurmans, Maarten Bosma, Brian Ichter, Fei Xia, Ed~Chi, Quoc~V Le, and Denny Zhou.
\newblock Chain-of-thought prompting elicits reasoning in large language models.
\newblock In \emph{Advances in Neural Information Processing Systems (NeurIPS)}, volume~35, pages 24824--24837, 2022.

\bibitem[Wu et~al.(2024)Wu, Zhao, Yasunaga, Huang, Cao, Huang, Ioannidis, Subbian, Zou, and Leskovec]{wu2024stark}
Shirley Wu, Shiyu Zhao, Michihiro Yasunaga, Kexin Huang, Kaidi Cao, Qian Huang, Vassilis~N. Ioannidis, Karthik Subbian, James Zou, and Jure Leskovec.
\newblock Stark: Benchmarking llm retrieval on textual and relational knowledge bases.
\newblock In \emph{Advances in Neural Information Processing Systems (NeurIPS) Datasets and Benchmarks Track}, 2024.
\newblock URL \url{https://arxiv.org/abs/2404.13207}.

\bibitem[Yu et~al.(2021)Yu, Liu, Xiong, Feng, and Liu]{yu2021fewshot}
Shi Yu, Zhenghao Liu, Chenyan Xiong, Tao Feng, and Zhiyuan Liu.
\newblock Few-shot conversational dense retrieval.
\newblock In \emph{Proceedings of the 44th International ACM SIGIR Conference on Research and Development in Information Retrieval}, pages 829--838, 2021.
\newblock \doi{10.1145/3404835.3462856}.

\bibitem[Zhang et~al.(2025)Zhang, Ai, Fan, Li, and Guo]{zhang2025fewshot}
Liang Zhang, Qingyao Ai, Yixing Fan, Haitao Li, and Jiafeng Guo.
\newblock Few-shot query intent detection via relation-aware prompt learning.
\newblock \emph{arXiv preprint arXiv:2509.05635}, 2025.

\bibitem[Zhu et~al.(2025)Zhu, Xie, Liu, Li, and Hu]{zhu2025kg2rag}
Xiangrong Zhu, Yuexiang Xie, Yi~Liu, Yaliang Li, and Wei Hu.
\newblock Knowledge graph-guided retrieval augmented generation.
\newblock In \emph{Proceedings of the 2025 Conference of the North American Chapter of the Association for Computational Linguistics (NAACL)}, 2025.
\newblock \doi{10.18653/v1/2025.naacl-long.449}.

\end{thebibliography}
\bibliographystyle{plainnat}

\end{document}